\def\year{2020}\relax
\documentclass[letterpaper]{article} % DO NOT CHANGE THIS
\usepackage{aaai20}  % DO NOT CHANGE THIS
\usepackage{times}  % DO NOT CHANGE THIS
\usepackage{helvet} % DO NOT CHANGE THIS
\usepackage{courier}  % DO NOT CHANGE THIS
\usepackage[hyphens]{url}  % DO NOT CHANGE THIS
\usepackage{graphicx} % DO NOT CHANGE THIS
\usepackage{graphicx}  % DO NOT CHANGE THIS
\usepackage{todonotes}
\usepackage{url}
\usepackage{amssymb}
\usepackage{array,multirow,makecell}
\usepackage{amsmath}
\usepackage{esvect}
\usepackage{bm}
\usepackage{diagbox}
\usepackage{changes}
\newcommand{\citet}[1]{\citeauthor{#1} \shortcite{#1}}

\title{Guiding attention in Sequence-to-sequence models for Dialogue Act prediction}

\author{ \textbf{Pierre Colombo\textsuperscript{\rm 1,2}\thanks{Equal contribution}, Emile Chapuis\textsuperscript{\rm 1}\footnote[1]{Equal contribution}, Matteo Manica\textsuperscript{\rm 3}}\\ \Large \textbf{Emmanuel Vignon\textsuperscript{\rm 2}, Giovana Varni\textsuperscript{\rm 1}, Chloe Clavel\textsuperscript{\rm 1}}\\
\textsuperscript{\rm 1}LTCI, Telecom Paris, Institut Polytechnique de Paris, \textsuperscript{\rm 2}IBM GBS France, \textsuperscript{\rm 3}IBM Zurich\\
\textsuperscript{\rm 1}firstname.lastname@telecom-paris.fr, \textsuperscript{\rm 3}tte@zurich.ibm.com, emmanuel.vignon@fr.ibm.com 
}

\begin{document}

\maketitle
\begin{abstract}

The task of predicting dialog acts (DA) based on conversational dialog is a key component in the development of conversational agents. Accurately predicting DAs requires a precise modeling of both the conversation and the global tag dependencies. We leverage seq2seq approaches widely adopted in Neural Machine Translation (NMT) to improve the modelling of tag sequentiality. Seq2seq models are known to learn complex global dependencies while currently proposed approaches using linear conditional random fields (CRF) only model local tag dependencies.  In this work, we introduce a seq2seq model tailored for DA classification using: a hierarchical encoder, a novel \textit{guided attention} mechanism and beam search applied to both training and inference. Compared to the state of the art our model does not require handcrafted features and is trained end-to-end. Furthermore, the proposed approach achieves an unmatched accuracy score of 85\% on SwDA, and state-of-the-art accuracy score of 91.6\% on MRDA.

\end{abstract}

\section{Introduction}
In natural language processing research, the dialogue act (DA) concept plays an important role. DAs are semantic labels associated with each utterance in a conversational dialogue that indicate the speaker's intention, e.g., question, backchannel, statement-non-opinion, statement opinion. A key to model dialogue is to detect the intent of the speaker: correctly identifying a question gives an important clue to produce an appropriate response.
\begin{table}[htb]
\begin{center}
\resizebox{\columnwidth}{!}{\begin{tabular}{c l}
Speaker & Utterance \\
 \hline
 A &Is there anyone who doesn't know Nancy?\\
 A &Do you - Do you know Nancy ?\\
 B &Me?\\
 B & Mm-hmm \\
 B& I know Nancy  \\
\end{tabular}}
\end{center}
\caption{Example of conversation from Switchboard Dialogue Act Corpus. A is speaking with B.}
\label{table:example1}
\end{table}
As can be observed in Table~\ref{table:example1}, DA classification relies on its conversational aspect, i.e., predicting an utterance's DA requires the knowledge of previous sentences and their associated act labels. For example, if a speaker asks a question, the interlocutor will answer with a response, analogously, a "Greeting" or a "Farwell" will be followed by a similar dialogue act. This means that in a conversation there is a sequential structure in the emitted dialogue acts. This poses the basis for the adoption of a novel perspective on the DA classification problem, i.e., from a multi-classification task to a sequence labeling one. 

\textbf{Limitations of current models:} Current state-of-the-art models rely on the use of linear Conditional Random Field (CRF) combined with a recurrent neural network based encoder \cite{crf_multi_task,LSTM_CRF,DAClassifCxt} to model DA sequential dependencies. Unfortunately such approaches only capture local dependencies between two adjacent dialogue acts. For instance, if we consider the example in Table~\ref{table:example1} we can see that the last statement "I know Nancy" is a response to the first question "Is there anyone who doesn't know Nancy" and the knowledge of the previous backchannel does not help the prediction of the last dialogue act. Therefore, we must consider dependencies between labels with a scope that is wider than two successive utterances. In Neural Machine Translation (NMT), the problem of global dependencies has been addressed using seq2seq models \cite{seq2seq} that follow the encoder-decoder framework. The encoder embeds an input sentence into a single hidden vector which contains both global and local dependencies, and the hidden vector is then decoded to produce an output sequence. In this work, we propose a seq2seq architecture tailored towards DA classification paving the way for further innovations inspired by advances in NMT research.

\textbf{Contributions:} In this work (1) we formalise the Dialogue Act Prediction problem in a way that emphasises the relations between DA classification and NMT, (2) we demonstrate that the seq2seq architecture suits better to the DA classification task and (3) we present a seq2seq model leveraging NMT techniques that reaches an accuracy of 85\%, outperforming the state of the art by a margin of around 2\%, on the Switchboard Dialogue Act Corpus (SwDA) \cite{switchboard_da} and a state-of-the-art accuracy score of 91,6\% on the Meeting Recorder Dialogue Act (MRDA).
This seq2seq model exploits a hierarchical encoder with a novel \textit{guided attention} mechanism that fits with our setting without any handcrafted features. We finetune our seq2seq using a sequence level training objective making use of the beam search algorithm. To our knowledge, this is among the first seq2seq model proposed for DA classification.

\section{Background}
\subsection{DA classification}
Several approaches have been proposed to tackle the DA classification problem. These methods can be divided into two different categories. The first class of methods relies on the independent classification of each utterance using various techniques, such as HMM \cite{handcrafted_HMM}, SVM \cite{svm_dialog} and Bayesian Network \cite{bayesian_dialog}. 
The second class, which achieves better performance, leverages the context, to improve the classifier performance by using deep learning approaches to capture contextual dependencies between input sentences \cite{RNN_CTX3_Softmax,LSTM_softmax}. Another refinement of input context-based classification is the modelling of inter-tag dependencies. This task is tackled as sequence-based classification where output tags are considered as a DA sequence \cite{crf_multi_task,hierarchical_LSTM_CRF,handcrafted_HMM,LSTM_CRF,DAClassifCxt}. 

Two classical benchmarks are adopted to evaluate DA classification systems: the Switchboard Dialogue Act Corpus (SwDA)\cite{switchboard_da} and the Meeting Recorder Dialogue Act (MRDA) \cite{mrda}. State-of-the-art techniques achieve an accuracy of 82.9\% \cite{crf_multi_task,DAClassifCxt}. To capture input contextual dependencies they adopt a hierarchical encoder and a CRF to model inter-tag dependencies. The main limitation of the aforementioned architecture is that a linear-CRF model is able to only capture dependencies at a local level and fails to capture non local dependencies. In this paper, we tackle this issue with a sequence-to-sequence using a guided attention mechanism.

\subsection{Seq2seq models}
Seq2seq models have been successfully applied to NMT, where modeling non local dependencies is a crucial challenge. DA classification can be seen as a problem where the goal is to map a sequence of utterances to a sequence of DA. Thus, it can be formulated as sequence to sequence problem very similar to NMT.\\
The general architecture of our seq2seq models \cite{seq2seq} follows a classical encoder-decoder approach with attention \cite{attention}. We use GRU cells \cite{grupaper}, since they are faster to train than LSTM ones~\cite{lstm_gru}. Recent advances have improved both the learning and the inference process, producing sequences that are more coherent by means of sequence level losses \citet{beam_seach_optimization} and various beam search settings~\cite{beam_search_alpha,dbs}. The closest setting where seq2seq model have been successfully used is dependency parsing \cite{dependancy}, where output dependencies are crucial to achieve state-of-the-art performance. In our work we adjust NMT techniques to the specifics of DA classification.

%Attention mechanism, let the seq2seq to focus on different part of the sequence each time a new word is generated and let the decoder to correctly align input sequence with output sequence. 
% by first training our seq2seq until converge using a simple token-level likelihood loss \cite{} and then fine tune the model by minimizing the Structured Prediction Losses.by first training our seq2seq until converge using a simple token-level likelihood loss \cite{} and then fine tune the model by minimizing the Structured Prediction Losses.

\section{Problem statement}
% \notette{This is redundant, I would avoid to speak about NMT at the beginning of the section. It's probably better to just say what we do. Something like: In this work we devised a seq2seq architecture for DA classification.}
% \deletedtte{
% To handle non local dependencies NMT have proposed several architecture. Seq2seq model are one of the most successful algorithm and currently reaches human level performances \cite{human_performances}.}
% In this work we devised a seq2seq architecture for DA classification.
\subsection{DA classification as an NMT problem}
First, let's define the mathematical notations we will adopt in this work. We have a set $D$ of conversations, i.e $D = (C_1,C_2,\dots,C_{|D|})$ with $Y= (Y_1,Y_2,\dots,Y_{|D|})$ the corresponding set DA labels. A conversation $C_i$ is a sequence of utterances, namely $C_i = (u_1,u_2,\dots,u_{|C_i|})$ with $Y_i = (y_1, y_2, \dots, y_{|C_i|})$ the corresponding sequence of DA labels. Thus, each utterance $u_i$ is associated with a unique DA label $y_i \in \mathcal{Y}$ where $\mathcal{Y}$ is the set of all the possible dialogue acts. Finally, an utterance $u_i$ can be seen as a sequence of words, i.e $u_i = (\omega^i_1, \omega^i_2, \dots, \omega^i_{|u_i|})$. In NMT, the goal is to associate for any sentence ${X}^{l_1} = (x^{l_1}_1, ..., x^{l_1}_{|{X}^{l_1}|})$ in language $l_{1}$ a sentence ${X}^{l_2} = (x^{l_2}_1, ..., x^{l_2}_{|{X}^{l_2}|})$ in language $l_{2}$ where $x^{l_k}_i$ is the $i$ word in the sentence in language $l_{k}$.
Using this formalism, it is straightforward to notice two main similarities ($\mathbb{S}_1$, $\mathbb{S}_2$) between DA classification and NMT. ($\mathbb{S}_1$) In NMT and DA classification, the goal is to maximise the likelihood of the output sequence given the input sequence ($P({X}^{l_2}|{X}^{l_1})$ versus $P({Y}_i|{C}_{i})$). ($\mathbb{S}_2$) For the two tasks, there are strong dependencies between units composing both the input and output sequences. In NMT, those units are words ($x_{i}$ and $y_{i}$), in DA classification those units are utterances and DA labels ($u_i$ and $y_i$).

\subsection{Specifics of DA classification}\label{ssec:differences}
While NMT and DA classification are similar under some point of views, three differences are immediately apparent ($\mathbb{D}_i$).
($\mathbb{D}_1$) In NMT, the input units $x_i$ represent words, in DA classification $u_i$ are input sequences composed with words. Considering the set of all possible sequences as input (context consideration leads to superior performance) implies that the dimension of the input space several order of magnitude larger than compared to a standard NMT. ($\mathbb{D}_2$) In DA, we have a perfect alignment between input and output sequences (hence $T =T\prime$). Some languages, e.g., French, English, Italian share a partial alignment, but in DA classification we have a strong mapping between $y_i$ and $x_i$. ($\mathbb{D}_3$) In NMT, the input space (number of words in $l_1$) is approximately the same size of the output space (number of words in $l_2$). In our case the output space (number of DA tags $|\mathcal{Y}| < 100$ has a limited size, with a dimension that is many order of magnitude smaller than the input space one.

In the following, we propose an end-to-end seq2seq architecture for DA classification that leverages ($\mathbb{D}_1$) using a hierarchical encoder, ($\mathbb{D}_2$) through a guided attention mechanism and ($\mathbb{D}_3$) using beam search during both training and inference, taking advantage of the limited dimension of the output space. 

\section{Models}
In Seq2seq, the encoder takes a sequence of sentences and represents it as a single vector $H_i \in \mathcal{R}^d$ and then pass it to the decoder for tag generations.
\begin{figure*}[!th]
   \center{\includegraphics[width=0.70\textwidth,origin=c]
       {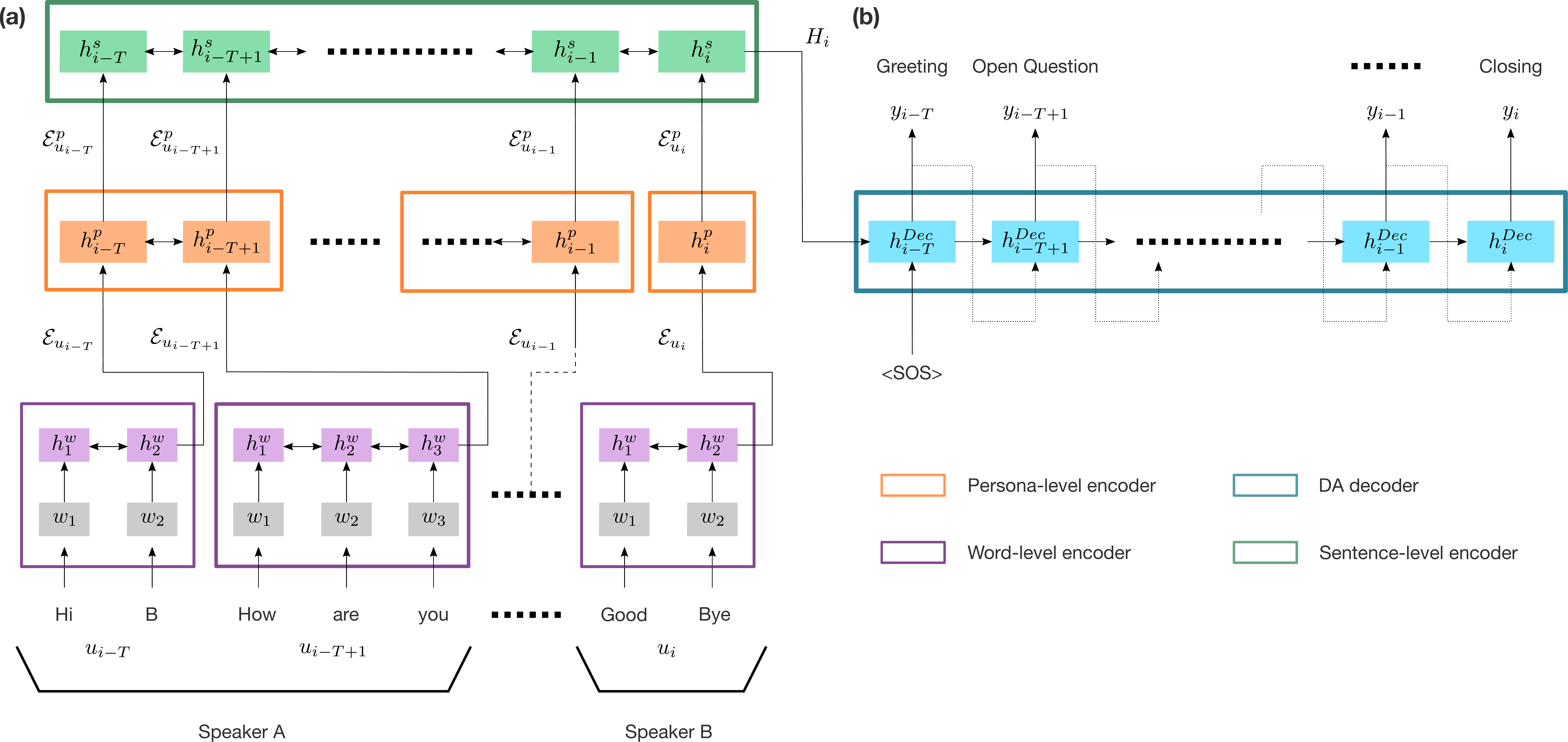}}
  \caption{\label{fig:architecture}Seq2seq model architecture for DA classification. (a) The encoder is composed with three different levels representing a different hierarchical level in the dialogue. The utterances are encoded at: word level (purple), persona level (orange) and sentence level (green). (b) The decoder (blue) is responsible to generate for each utterance a DA exploiting the last state of the encoder as initial hidden state.
   } 
\end{figure*}

\subsection{Encoders}
In this section we introduce the different encoders we consider in our experiments. We exploit the hierarchical structure of the dialogue to reduce the input space size ($\mathbb{D}_{1}$) and to preserve word/sentence structure. During both training and inference, the context size is fixed to $T$. Formally, an encoder takes as input a fixed number of utterances ($u_{i-T},..,u_{i}$) and outputs a vector $H_i \in \mathcal{R}^d$ which will serve to initialize the hidden state of the decoder. The first level of the encoder computes $\mathcal{E}_{u_t}$, an embedding of $u_t$ based on the words composing the utterance, and the next levels compute $H_i$ based on $\mathcal{E}_{u_t}$.

\noindent\textbf{Vanilla RNN encoder:} The vanilla RNN encoder ($\text{VGRU}_\text{E}$) introduced by \citet{seq2seq} is considered as a baseline encoder. In the vanilla encoder $\mathcal{E}_{u_i} = \frac{1}{|u_i|} \sum_{k=1}^{|u_i|} \mathcal{E}_{w^i_k}$ where $\mathcal{E}_{w^i_k}$ is an embedding of $w^i_k$. To better model dependencies between consecutive utterances, we use a bidirectional GRU \cite{grupaper}:

\begin{equation} \label{eq:gru_layer_ht}
 \begin{array}{l}
 \overrightarrow{h_{i-T}^s} = \overleftarrow{h_{i-T}^s} = \overrightarrow{0} \\

    \overrightarrow{h^s_t} = \overrightarrow{GRU}(\mathcal{E}_{u_t}), t \in [i-T,i]  \\
     \overleftarrow{h^s_t} = \overleftarrow{GRU}(\mathcal{E}_{u_t}), t \in [i,i-T] \\

  H_{i} = [\overleftarrow{h^s_{i}},\overrightarrow{h^s_{i}}]
 
   \end{array}
\end{equation}

\noindent\textbf{Hierarchical encoders:} The vanilla encoder can be improved by computing $\mathcal{E}_{u_i}$ using bi-GRU. This hierarchical encoder (HGRU) is in line with the one introduced by \citet{hierarchical_encoder}. Formally $\mathcal{E}_{u_i}$ is defined as it follows:
\begin{equation}\label{eq:gru_layer_word}
 \begin{array}{l}
 \overrightarrow{h^w_0} = \overleftarrow{h^w_0} = \overrightarrow{0} \\
  \overrightarrow{h^w_t} = \overrightarrow{GRU}(\mathcal{E}_{w^i_t}), t \in [1,|u_i|]\\
 \overleftarrow{h^w_t} = \overleftarrow{GRU}(\mathcal{E}_{w^i_t}), t \in [|u_i|,1]\\
  \mathcal{E}_{u_i}= [\overleftarrow{h^w_{|u_i|}},\overrightarrow{h^w_{|u_i|}}]
   \end{array}
\end{equation}
$H_{i}$ is then computed using Equation~\ref{eq:gru_layer_ht}. Intuitively, the first GRU layer (Equation~\ref{eq:gru_layer_word}) models dependencies between words (the hidden state of the word-level GRU is reset at each new utterance), and the second layer models dependencies between utterances.

\noindent\textbf{Persona hierarchical encoders:} In SwDA, a speaker turn can be splitted in several utterances. For example, if speaker A is interacting with speaker B we might encounter the sequence (AAABBBAA)\footnote{In SwDA arround two third of the sentence have at least a AA or BB}. We propose a novel Persona Hierarchical encoder (PersoHGRU) to better model speaker-utterance dependencies. We introduce a persona layer between the word and the sentence levels, see Figure~\ref{fig:architecture}:
\begin{equation} \label{eq:gru_layer_persona}
 \begin{array}{l}
    %\overrightarrow{h_{i-T}^{ps}} = \overleftarrow{h_{i-T}^s} = \overrightarrow{0} \\
    \overrightarrow{h_{t}^{p}} = 
    \left \{
   \begin{array}{l}
      \overrightarrow{0} \text{ if $t$ and $t-1$ have different speakers} \\
      \overrightarrow{GRU}(\mathcal{E}_{u_{t-1}})
   \end{array}
    \right .
    \\
       \overleftarrow{h_{t}^{p}} = 
    \left \{
   \begin{array}{l}
      \overrightarrow{0} \text{ if $t$ and $t+1$ have different speakers} \\
      \overleftarrow{GRU}(\mathcal{E}_{u_{t+1}})
   \end{array}
    \right .
    \\
    \mathcal{E}^p_{u_k} = [\overrightarrow{h_{k}^{p}},\overleftarrow{h_{k}^{p}}]\hspace{.5cm} \forall k \in [i-T,i]
   \end{array}
\end{equation}
$H_{i}$ is then obtained following Equation~\ref{eq:gru_layer_ht} where $\mathcal{E}_{u_i}$ is replaced by $\mathcal{E}^p_{u_i}$.

\subsection{Decoders}
In this section, we introduce the different decoders we compare in our experiments. We introduce a novel form of attention that we name \textit{guided attention}. \textit{Guided attention} leverages the perfect alignment between input and output sequences ($\mathbb{D}_2$). The decoder computes the probability of the sequence of output tags based on:
\begin{small}
\begin{equation}
 p(y_{i-T},\dots,y_{i}| u_{i-T}, …, u_i) = \prod_{k=i-T}^i  p(y_k|H_i,y_{k-1},\dots,y_{i-T})
\end{equation}
\end{small}
see Equation~\ref{fig:architecture}.

\indent\textbf{Vanilla decoder:} The vanilla decoder ($\text{VGRU}_\text{D}$) is similar to the one introduced by \citet{seq2seq}.

\indent\textbf{Decoders with attention:} In NMT, the attention mechanism forces the seq2seq model to learn to focus on specific parts of the sequence each time a new word is generated and let the decoder correctly align the input sequence with output sequence. In our case, we follow the approach described by \citet{badau} and we define the context vector as:
\begin{equation}
\begin{array}{c}
     c_k = \displaystyle\sum_{j=i-T}^i \alpha_{j,k}  h_{j}^s
\end{array}
\end{equation}
where $ \alpha_{j,k}$  scores how well the inputs around position $k$ and the output at position $j$ match. Since we have a perfect alignment ($\mathbb{D}_2$), we know a priori on which sequence the decoder needs to focus more at each time step. Taking into account this aspect of the problem, we propose three different attention mechanisms.

\noindent\textbf{Vanilla attention:} This attention represents our baseline attention mechanism and it is the one proposed by \citet{badau}, where: \begin{equation}
     \alpha_{j,k} = softmax(a(h_{k-1}^{Dec}, h_{j}^s))
\end{equation} and $a$ is parametrized as a feedforward neural network.

\noindent\textbf{Hard guided attention:} The \textit{hard guided attention} forces the decoder to focus only on the $u_{i}$ while predicting $y_i$:
\begin{equation}
        \alpha_{j,k} = \begin{cases}
    0,& \text{if } k \ne j \\
    1 ,              & \text{otherwise}
    \end{cases}
\end{equation}

\indent\textbf{Soft guided attention:} The \textit{soft guided attention} guides the decoder to mainly focus on the $u_{i}$ while predicting $y_i$, but allows it to have a limited focus on other parts of the input sequence. 
\begin{align}
    \widetilde{\alpha}_{j,k} &= \begin{cases}
    a(h_{k-1}^{Dec}, h_{j}^s) ,& \text{if } k \ne j \\
    1 + a(h_{k-1}^{Dec}, h_{j}^s),              & \text{otherwise}
    \end{cases}\\
    \alpha_{j,k} &= softmax(\widetilde{\alpha}_{j,k})
\end{align}
where $a$ is parametrised  as a feedforward neural network.

\subsection{Training and inference}\label{secc:trainAndInfer}
In this section, we describe the training and the inference strategies used for our models. A seq2seq model aims to find the best sentence for a given source sentence. This poses a computational challenge when the output vocabulary size is large, since even by using beam search it's expensive to explore multiple paths. Since our output vocabulary size is limited ($\mathbb{D}_3$), we do not incur in this problem and we can use beam search during both training and inference.

\noindent\textbf{Beam search:} In our work we measure the sequence likelihood based on the following formula:
\begin{equation}
    s(\tilde{\bm{y}}^k,\bm{u}_i)=\frac{\log P(\tilde{\bm{y}}^k|\bm{u}_i) }{lp(\tilde{\bm{y}}^k)}
\end{equation}
where $\bm{u}_i = (u_{i-T},\dots, u_i)$ and $\tilde{\bm{y}}^k = (\tilde{y}_{i-T},\dots, \tilde{y}_{i-T+k})$ is the current target, and $lp(\tilde{\bm{y}})=\frac{(5+|\tilde{\bm{y}}|)^\alpha}{(5+1)^\alpha}$ is the length normalisation coefficient \cite{beam_search_alpha}. At each time step the $B$ most likely sequences are kept ($B$ corresponding to the beam size).

\noindent\textbf{Training objective:} For training we follow \citet{beam_seach_optimization} and train our model until convergence with a token level loss and fine tune it by minimising the expected risk $\mathcal{L}_\text{RISK}$ defined as:
\begin{equation}
    \mathcal{L}_\text{RISK} = \displaystyle\sum_{\tilde{\bm{y}} \in \text{U}(C_{i})} \dfrac{cost(\tilde{\bm{y}},\bm{y}_i) p(\tilde{\bm{y}} | \bm{u}_i)}{ \displaystyle\sum_{\tilde{\bm{y}}' \in \text{U}(\bm{u}_i)}  p(\tilde{\bm{y}}| \bm{u}_i)}
\end{equation}
where $U(\bm{u}_i)$ is the set of the sequences generated by the model using a beam search algorithm for the input $\bm{u}_i$, and $cost(\tilde{\bm{y}},\bm{y}_i)$ is defined, for a given a candidate sequence
$\tilde{\bm{y}}$ and a target $\bm{y}_i$, as:
\begin{equation}
        cost(\tilde{\bm{y}},\bm{y}_i) = \begin{cases}
    1& \text{if } \tilde{\bm{y}}_{i} = \bm{y}_{i} \\
    0              & \text{otherwise}
    \end{cases}
\end{equation}

\subsection{GRU/HGRU CRF baseline}
State-of-the-art models use conditional random fields which model dependencies between tags on top of an GRU or a HGRU encoder which computed an embedding of the a variable number of utterances sentences .
%For dialogue act classification, in the setting where a variable number of utterances can be processed as input, state-of-the-art models \added{feed embedding of the input sentences computed using a GRU or a HGRU encoder to  conditional random fields which model dependencies between tags.} 
We have implemented our own CRF ($\text{Baseline}_{\text{CRF}}$) following the work of \citet{hierarchical_LSTM_CRF}:
\begin{equation}
 \begin{array}{l}
        p(y_i,\dots,y_{i-T},u_i,\dots, u_{i-T};\theta) =\\ \frac{\displaystyle\prod_{t=i}^{T}\psi(y_{t-1},y_{t},\phi(o_t); \theta)}{\displaystyle\sum_{\mathcal{Y}} {\displaystyle\prod_{t=i}^T}\psi(y_{t},y_{t-1},\phi(o_t); \theta)}
        \end{array}
\end{equation}
Here $\theta$ is the set of parameters corresponding to the CRF layer, and $\psi$ is the feature function, providing us with unary and pairwise potentials. Let $\phi \colon \mathbb{R}^H \to \mathbb{R}^{|\mathcal{Y}|}$ be the dense representation of each utterance's output provided by the encoder. $\phi$ can be seen as the unary feature function.

\section{Experimental Protocol}
In this section we describe the experimental protocols adopted for the evaluation of our approach.

\subsection{Datasets} \label{ssec:datasets}
%\addedspeak about spoken language that this phenomenon is very common

We consider two classical datasets for Dialogue Act Classification: The Switchboard Dialogue Act Corpus and the MRDA. Since our models explicitly generate a sequence of tags we compute the accuracy on the last generated tag.
Both datasets are already segmented in utterances and each utterance is segmented in words.
For each dataset, we split each conversation $C_{i}$ in sequence of utterances of length $T=5$\footnote{$T$ is an hyperparameter, experiments have shown that 5 leads to the best results.}.

\noindent\textbf{SwDA:} The Switchboard-1 corpus is a telephone speech corpus \cite{switchboard_da}, consisting of about 2.400 two-sided telephone conversation among 543 speakers with about 70 provided conversation topics. The dataset includes information about the speakers and the topics and has 42 different tags.
In this dataset global dependency plays a key role due to the large amount of backchannel (19\%), abandoned or turn-exit (5\%), uninterpretable acts (1\%). In this context, any models that only take into account local dependencies will fail at extracting information to distinguish between ambiguous tags. 
For the confusion matrix, we follow \citet{crf_multi_task} and present it for 10 tags only: statement-non-opinion (\texttt{sd}), backchannel (\texttt{b}), statement-opinion (\texttt{sv}), conventional-closing (\texttt{fc}), wh-question (\texttt{qw}), response acknowledgement (\texttt{bk}), hedge (\texttt{h}), open-question (\texttt{qo}), other answers (\texttt{no}), thanking (\texttt{ft}).

\noindent\textbf{MRDA:} MRDA: The ICSI Meeting Recorder Dialogue Act corpus \cite{mrda_2} contains 72 hours of naturally occurring multi-party meetings that were first converted into 75 word level conversations, and then hand-annotated with DAs using the Meeting Recorder Dialogue Act Tagset. In this work we use 5 DAs, i.e., statements (\texttt{s}), questions (\texttt{q}), floorgrabber (\texttt{f}), backchannel (\texttt{b}), disruption (\texttt{d}).

\noindent\textbf{Train/Dev/Test Splits}: For both SwDA and RMDA we follow the official split introduced by \citet{handcrafted_HMM}. Thus, our model can directly be compared to \citet{crf_multi_task,LSTM_CRF,hierarchical_LSTM_CRF,DAClassifCxt}.

\subsection{Training details}
All the hyper-parameters have been optimised on the validation set using accuracy computed on the last tag of the sequence. The embedding layer is initialised with pretrained fastText word vectors of size 300 \cite{fastText}\footnote{In our work we rely on same pretrained embedding word2vect \cite{word2vec} instead of GloVe \cite{glove}.} , trained with subword information (on Wikipedia 2017, UMBC webbase corpus and statmt.org news dataset), and updated during training. Hyperparameter selection has been done using a random search on a fixed grid.
Models have been implemented in PyTorch and trained on a single NVIDIA P100.

\noindent\textbf{Parameters for SwDA:} We used Adam optimizer \cite{adam} with a learning rate of 0.01, which is updated using a scheduler with a patience of 20 epochs and a decrease rate of 0.5. The gradient norm is clipped to 5.0, weight decay is set to 1e−5, and dropout \cite{dropout} is set to 0.2. The maximum sequence length is set to 20.  Best performing model is an encoder with size of 128 and a decoder of size 48.  For $\text{VGRU}_\text{E}$, we use two layers for the BiGRU layer. For hierarchical models, we use BiGRU with a single layer.

\noindent\textbf{Parameters for MRDA:} We used AdamW optimizer \cite{adamw} with a learning rate of 0.001, which is updated using a scheduler with a patience of 15 epochs and a decrease rate of 0.5. The gradient norm is clipped to 5.0, weight decay is set to 5e−5, and dropout \cite{dropout} is set to 0.3. The maximum sequence length is set to 30.  Best performing model is an encoder with size of 40 and a decoder with size 400.  For $\text{VGRU}_\text{E}$ we use two layers for the BiGRU layer, for hierarchical models we use BiGRU with a single layer.

% \subsection{Evaluation criteria}
% Classes are widely imbalance so the F1 score seems to be rather than the accuracy a more suitable metric to evaluate our models. %More precisely we only evaluate the last sentence of the context. 
% Given the testing conversation $C$ a set a $N$ sentences i.e $C = \{s_1,\dots,s_N\}$ and the associated dialogue act label $Y =\{y_1,\dots,y_N\}$. 

% \begin{equation}
%     F_1_{\text{micro}} = \frac{1}{N}\sum_{i=1}^N \mathbb{1}\{\hat{y}_i \neq y_i\}
% \end{equation}

\section{Experiments \& Results}
In this section we propose a set of experiments in order to investigate the performance of our model compared to existing approaches with respect to the difficulties highlighted in the introduction. 
\subsection{Experiment 1: Are Seq2seq better suited to DA prediction than CRF ?}
Current state of the art are built on CRF models. In this first section, we aim at comparing a seq2seq with a CRF based model. To provide a fair comparison we perform the same grid search for all models on a fixed grid. At this step, we do not use attention neither use beam search during training or inference. As shown in Table~\ref{tab:seq_vs_crf},  with a vanilla RNN encoder the seq2seq significantly outperforms the CRF on SwDa and MRDA. With an HGRU the seq2seq exhibit significantly higher results on SwDA and reaches comparable performances on MRDA. This behaviour suggests that a model based on a seq2seq architecture tends to be achieve higher score on DA classification than a CRF based model. 

\begin{table}[!htb]
\centering
\begin{tabular}{c|c|c}
\hline
    Models & SwDa & MRDA\\ 
    \hline
         $\text{Baseline}_\text{CRF}$ (+GRU)  & 77.7 & 88.3\\ 
     seq2seq (+GRU) & \textbf{81.9}& \textbf{88.5}\\

    \hline
      $\text{Baseline}_\text{CRF}$ (+HGRU)   & 81.6 & 90.0\\ 
     seq2seq (+HGRU) & \textbf{82.4}&90.0\\

     \hline
\end{tabular}
\caption{Accuracy of a seq2seq on dev test and $\text{Baseline}_\text{CRF}$ on SwDA and MRDA. Bold results exhibit significant differences ($\text{p-value} < 0.01$) according to the Wilcoxon Mann Whitney test performed on 10 runs using different seeds. }
\label{tab:seq_vs_crf}
\end{table}

\noindent\textbf{Global dependencies analysis:} In Table~\ref{tab:comp_ex} we present two examples where our seq2seq use contextual information to disambiguate the tag and to predict the correct label. In the first example, ``It can be a pain" without context can be interpreted both as statement non-opinion (\texttt{sd}) or statement opinion (\texttt{sv}). Our seq2seq uses the surrounding context (two sentences before) to disambiguate and assign the \texttt{sv} label . In the second example, the correct tag assigned to ``Oh, okay'' is a response acknowledgement (\texttt{bk}) and not backchannel (\texttt{b}). The key difference between \texttt{bk} and \texttt{b} is that an utterance labelled with \texttt{bk} has to be produced within a question-answer context, whereas \texttt{b} is a \textit{continuer}
\footnote{This analysis can be supported by 5.1.1 in SwDA coder manual \url{https://web.stanford.edu/~jurafsky/ws97/manual.august1.html}}.
In our example, the global context this is a question/reply situation: the first speaker asks a question (``What school is it''), the second replies then, the first speaker answers to the reply. This observation reflects the fact CRF models only handle local dependencies where seq2seq models consider global ones as well.  

\begin{table}[!htb]\label{tab:kysymys}
\centering
\resizebox{\columnwidth}{!}{\begin{tabular}{l|ccc} 
\hline
Utterances & G. & $\text{seq2seq}$  & CRF \\
\hline
How long does that take you to get to work? &	\texttt{qw}& \texttt{qw}& \texttt{qw} \\
Uh, about forty-five, fifty minutes.&	\texttt{sd}& \texttt{sd}& \texttt{sd} \\
\makecell[l]{How does that work, work out with, uh,\\storing your bike and showering and all that?}
&	\texttt{qw}& \texttt{qw}& \texttt{qw}  \\
Yeah , &	\texttt{b} &\texttt{b} &\texttt{b}\\
It can be a pain .	&\texttt{sd}& \texttt{sd}& \texttt{sv}\\
\hline
\makecell[l]{It's, it's nice riding to school because\\ it's all along a canal path, uh,}&	\texttt{sd} &\texttt{sd} &\texttt{sd} \\ 
\makecell[l]{Because it's just, \\it's along the Erie Canal up here.}&	\texttt{sd} &\texttt{sd} &\texttt{sd} \\ 
So, what school is it?&	\texttt{qw} &\texttt{qw} &\texttt{qw} \\ 
Uh, University of Rochester.&	\texttt{sd}& \texttt{sd} &\texttt{sd} \\ 
Oh, okay.&	\texttt{bk}& \texttt{bk} &\texttt{b} \\ 
\hline
\end{tabular}}
\caption{Example of predicted sequence of tags taken from SwDA. $\text{seq2seq}$ is our best performing model, CRF stands for $\text{Baseline}_\text{CRF}$, G. is the groundtruth label.}
\label{tab:comp_ex}
\end{table}

\begin{table}[!htb]

\centering
\begin{tabular}{c|cccc}
\hline 
&\multicolumn{4}{c}{SwDA} \\ \hline
     \diagbox[height=20pt]{\footnotesize{Enc.}}{\footnotesize{Dec.}}  & \multicolumn{1}{c}{$\text{VGRU}_\text{D}$} & \multicolumn{1}{c}{att.} & \multicolumn{1}{c}{soft guid.} & \multicolumn{1}{c}{hard guid.}\\ \hline
     Beam Size &  1    &1  & 1 & 1 \\ \hline

   {$\text{VGRU}_\text{E}$}  & 81.6   &82.1  &   82.8   &82.9 \\ 
   {HGRU} & 82.4    &82.3    &   83.1 & \textbf{84.0} \\ 
   {PersoHGRU}  & 49.8     &79.4    &   84.0 &83.5 \\ \hline
    
&\multicolumn{4}{c}{MRDA} \\\hline
      \diagbox[height=20pt]{\footnotesize{Enc.}}{\footnotesize{Dec.}} & \multicolumn{1}{c}{$\text{VGRU}_\text{D}$} & \multicolumn{1}{c}{att.} & \multicolumn{1}{c}{soft guid.} & \multicolumn{1}{c}{hard guid.} \\ \hline
     Beam Size &  1   &1  & 1 &1\\ \hline

   {$\text{VGRU}_\text{E}$}  & 88.5     & 88.5&88.5&88.5    \\ 
   {HGRU} & 90.0    &89.9     &   90.0&\textbf{90.2} \\ 
   {PersoHGRU} & 66.2     &87.7     &   88.2  &86.9     \\ 
   
     \hline
\end{tabular}
\caption{Accuracy on the dev set of the different encoder/decoder combination  MRDA and SwDA. For SwDA, Wilcoxon test (10 runs with different seeds) has been performed for an HGRU encoder with a decoder with \textit{hard guided attention} against  an HGRU encoder with \textit{soft guided attention}, \textit{soft guided attention}, with attention, without attention pairwise tests exhibit $\text{p-value} < 0.01$.}
\label{tab:f1_all}
\end{table}

\subsection{Experiment 2: What is the best encoder?}
%\subsubsection{Encoder}
In Table~\ref{tab:f1_all}, we present the results of the three encoders presented in Section 4 on both datasets. For SwDA and MRDA, we observe that a seq2seq equipped with a hierarchical encoder outperforms models with Vanilla RNN encoder, while reducing the number of learned parameters.

The $\text{VGRU}_\text{D}$ does not play well with the PersoHGRU encoder. When combined with a \textit{guided attention mechanism}, the PersoHGRU exhibits competitive accuracy on SwDA. However on MRDA, adding a personna layer harms the accuracy. This suggests either that the information related to the speaker is irrelevant for our task (no improvement observed while adding persona information) \footnote{Further investigations with several persona based model inspired from the work of \cite{persona_based} shows the same poor improvement (in terms of accuracy)}, or that the considered hierarchy is not the optimal structure to leverage this information. 
%\added{On SwDA the proportion of speaker changes at each turn is $0.56$ vs $0.63$ for MRDA (a relative difference of 12\%). In other words, SwDA speakers alternate less often than in MRDA which explains the difference in the performance of the PersoHGRU on MRDA and SwDA.} 

Our final model makes use of the HGRU encoder since in most of the settings it exhibits superior performance.

%It demonstrates that alignment information is crucial for the learning process of our seq2seq model. 
%On SwDA, with an encoder using a \textit{soft guided attention} mechanism we observe a relatively small improvement (0.2\%) of the PersoHGRU over the HGRU. This suggests either that the information relative to the speaker is relevant for our task\footnote{Further investigations with several persona based model inspired from the work of \cite{persona_based} shows the same poor improvement (in terms of accuracy)}, or that the considered hierarchy is not the optimal structure to leverage this information.

\subsection{Experiment 3: Which attention mechanism to use?}
%\ref{tab:f1_all} shows that the $\text{VGRU}_\text{D}$ achieves competitive performance testifying the quality of the encoded representation, $H_{i}$.

The seq2seq encodes a source sentence into a fixed-length vector from which a decoder generates a sequence of tags. Attention forces the decoder to strengthen its focus on the source sentences that are relevant to predicting a label.

In NMT \cite{attention}, complementing a seq2seq with attention contributes to generate better sentences. In Table~\ref{tab:f1_all} we see that in most the case, the use of a simple attention mechanism provides a rather small improvement with VGRU and harms a bit the performances with a HGRU encoder. In case of a seq2seq composed with a PersoHGRU and a decoder without attention the learning fails: the decrease of the training loss is relatively small and seq2seq fails to generalise. It appears that in DA classification where sequences are short (5 tags), Vanilla attention does not have as much as impact as in NMT (that have longer sequences with more complex global dependencies). 

If we consider an HGRU encoder, we observe that our proposed \textit{guided attention} mechanisms improves dev accuracy which demonstrates the importance of easing the task by using prior knowledge on the alignment between the utterances and the tags. Indeed, while decoding there is a direct correspondence between labels and utterances meaning that $y_i$ is associated with $u_i$. The soft guided attention will mainly focus on the current utterance with a small additional focus on the context where hard guided attention will only consider the current utterance. Improvement due to \textit{guided attention} demonstrates that the alignment between input/output is a key prior to include in our model.

\begin{figure*}[!htb]
\centering
  \includegraphics[width=0.70\textwidth]{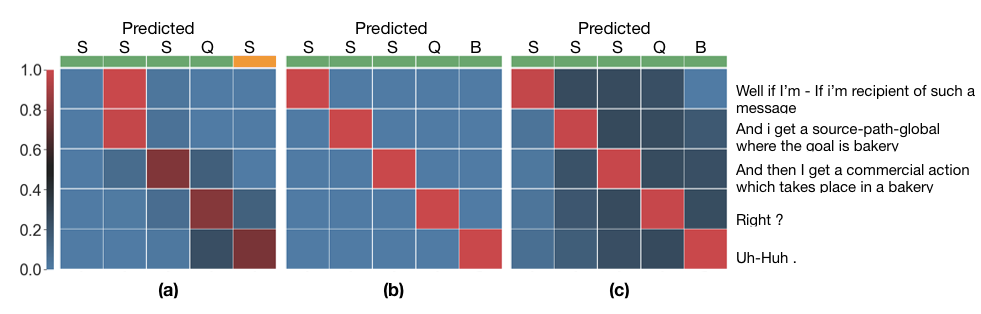}
\caption{Attention matrix visualisation on MRDA for the fixed context of 5 utterances. Green color for predicted label indicates a correct label, orange color indicates a mistake. (a) stands for the HGRU with attention, (b) stands for the HGRU with hard guided attention, (c) is HGRU with soft guided attention. }
\label{fig:attention_comparison}
\end{figure*}

\noindent\textbf{Attention analysis:} Figure~\ref{fig:attention_comparison} shows a representative example of the attention weights of the three different mechanisms. The seq2seq with a normal attention mechanism is characterised by a weight matrix far from the identity (especially the lower right part). While decoding the last tags, this lack of focus leads to a wrongly predicted label for a simple utterance: ``Uh-Huh'' (backchannel). Both \textit{guided attention} mechanisms focus more on the sentence associated with the tag, at each time step, and predict successfully the last DA.

Since the \textit{hard guided attention} decoder exhibit overall the best results (on both SwDA and MRDA) and does not require any additional parameter we will use it for our final model.

\subsection{Experiment 4: How to leverage beam search to improve the performance?}
Beam Search allows the seq2seq model to consider alternative paths in the decoding phase.

\noindent\textbf{Beam Search during inference:} Using beam search provides a low improvement (maximum absolute improvement of $0.2\%$) %Full results are provided in Appendix Table \ref{tab:f1_all_all}
\footnote{The considered beam size are small compared to other applications \cite{mmi}. While increasing the beam size, we see that the beam search become very conservative~\cite{diversity_bea_search} and tends to output labels highly represented in the training set (e.g., \texttt{sd} for SwDA).}.

Compared to NMT, output size is drastically smaller  ($\mathcal{Y}_\text{SwDA} = 42$ while $\mathcal{Y}_\text{MRDA} = 5$) for DA classification. When considering alternative paths with small output space in imbalanced datasets the beam search is more likely to consider very unlikely sequences as alternatives (eg. ``\texttt{s} \texttt{s} \texttt{s} \texttt{s} \texttt{s}''). 

\noindent\textbf{Fine tuning with a sequence loss:} As previously mentioned, using beam search during inference only leads to a limited improvement in accuracy. We finetune a seq2seq composed with a HGRU encoder and a decoder with \textit{hard guided attention} (this model has been selected in the previous steps) with the introduced sequence level loss describes in Section~\ref{secc:trainAndInfer}. 
Table~\ref{tab:swa_finetuning} shows that this fine tuning steps improves the performances of  1\% on SwDA (84\% vs 85\%) and 1.2\% on RMDA (90.4\% vs 91.6\%). 

\begin{table}[!htb]

\centering
\begin{tabular}{c|cc|cc}
\hline 
  &\multicolumn{2}{c}{SwDA} \vline &\multicolumn{2}{c}{RMDA} \\\hline
    \diagbox[height=20pt]{\footnotesize{$B_\text{inf}$}}{\footnotesize{$B_\text{train}$}} 
    & 2 & 5& 2 & 5 \\ 
    \hline
   {1}  & 84.8 & 84.7& 91.3 & \textbf{91.6}\\ 
   {2} & 84.9 & 84.8& 91.3 & 91.6\\ 
   {5}  & \textbf{85.0} & 84.9& 91.5 & 91.6\\ 
     \hline
\end{tabular}
\caption{Accuracy on the dev set of seq2seq model trained with sequence level loss. $B_\text{train}$ stands for the beam size during training, $B_\text{inf}$ for the one during inference\protect\footnotemark.For SwDA, Wilcoxon test (10 runs with different seeds) has been performed for $B_\text{train}=2$ and $B_\text{inf}=2$ against all other models. For RMDA, Wilcoxon test has been performed (10 runs with different seeds) for $B_\text{train}=5$ and $B_\text{inf}=1$ against all model with  $B_\text{train}=2$.}

\label{tab:swa_finetuning}
\end{table}

\noindent\textbf{$\text{seq2seq}_\text{BEST}$:} Our $seq2seq_{BEST}$ model is composed of a HGRU encoder and a decoder with \textit{hard guided attention} finetuned with $B_\text{train} = 2$ and $B_\text{inf} = 5$ for SwDA and $B_\text{train} = 5$ and $B_\text{inf} = 1$ for SwDA.

%\subsection{Performances Analysis}
\subsection{Experiment 5: Comparison with state-of-the-art models}
In this section, we compare the performances of $seq2seq_{BEST}$ with other state of the art models and analyse the performances of the models.
Table~\ref{tab:state_of_the_art} shows the performances of best performing model $\text{seq2seq}_\text{BEST}$ on the test set. $\text{seq2seq}_\text{BEST}$ achieves an accuracy of 85\% on the SwDA corpora.  This model outperforms \citet{LSTM_CRF} and \citet{DAClassifCxt} which achieve an accuracy of 82.9\%. On MRDA, our best performing model reaches an accuracy of 91.6\% where current state-of-the-art systems, \citet{LSTM_CRF,hierarchical_LSTM_CRF} achieve respectively 92.2\% and 91.7\%.
%and \citet{hierarchical_LSTM_CRF}\added{and \citet{DAClassifCxt}}, achieve respectively 91.7\% and 90.9\% \added{92.2\% and 91.1\%}.

\begin{table}[!htb]
\centering
\begin{tabular}{c|c|c}
\hline
    Models & SwDa & MRDA\\ 
    \hline
   \citet{crf_multi_task}& 82.9 & 92.2 \\
   %\citet{LSTM_CRF} & \added{82.3}81.3& 91.7\added{}  \\
   \citet{LSTM_CRF} & 81.3 & 91.7  \\
  \citet{hierarchical_LSTM_CRF} &79.2& 90.9 \\ 
  \citet{DAClassifCxt} & 82.9 & 91.1 \\
       \hline
    $\text{seq2seq}_\text{BEST}$ & \textit{85.0}& \textit{91.6}\\
     \hline
\end{tabular}
\caption{Accuracy of our best models (seq2seq) and $\text{Baseline}_\text{CRF}$ on SwDA and MRDA test sets.}
\label{tab:state_of_the_art}
\end{table}

\section{Conclusion}
In this work, we have presented a novel approach to the DA classification problem. We have shown that our seq2seq model, using a newly devised \textit{guided attention} mechanisms, achieves state-of-the-art results thanking its ability to better model global dependencies.\\
%\added{Future work will include Transformer like architecture.}

\section*{Acknowledgement}
This work was supported by a grant overseen from the French National Research Agency (ANR-17-MAOI).
%Our model could be improve with the most recent advances in NMT: the adoption of more complex seq2seq models (e.g ConvS2S \cite{cov2s}, Transformer~\cite{transformer,bert}) or the implementation of a advanced beam search strategy \cite{dbs}.

%Our formulation offers multiple possibilities for improvements.
%These improvements could leverage: the consideration of the context ($\mathbb{D}_1$), the perfect alignment between utterances and tags ($\mathbb{D}_2$), and the limited size of the vocabulary for the decoded sequence ($\mathbb{D}_3$).
%For example, f

%and would make a better use of the limited size of the output space.
%Being able to predict DA tags using contextual dependencies and to handle dialogues of variable length can greatly help to generate natural language in a dialogue-act-driven fashion, leading to consistent improvements to the quality of conversational agents.

% Another line of research could include adapting a seq2seq problem to similar tasks such as sentiment, opinion analysis in a conversation. 

% what about conversational agent? you should re-take the stuff in the introduction...
% The conclusions have to be expanded and completed. The conversational agent should be brought up again I agree. I can go through them once they are here.

% \noindent \textbf{Preparing References:} \\
% Include your own bib file like this:
%\verb|\bibliographystyle{acl_natbib}|
% \verb|\bibliography{acl2019}| 

% where \verb|acl2019| corresponds to a acl2019.bib file.

\bibliography{biblio}

\begin{thebibliography}{}

\bibitem[\protect\citeauthoryear{Bahdanau, Cho, and Bengio}{2014}]{badau}
Bahdanau, D.; Cho, K.; and Bengio, Y.
\newblock 2014.
\newblock Neural machine translation by jointly learning to align and
  translate.
\newblock {\em CoRR} abs/1409.0473.

\bibitem[\protect\citeauthoryear{Bojanowski \bgroup et al\mbox.\egroup
  }{2017}]{fastText}
Bojanowski, P.; Grave, E.; Joulin, A.; and Mikolov, T.
\newblock 2017.
\newblock Enriching word vectors with subword information.
\newblock {\em Transactions of ACL} 5:135--146.

\bibitem[\protect\citeauthoryear{Bothe \bgroup et al\mbox.\egroup
  }{2018}]{RNN_CTX3_Softmax}
Bothe, C.; Weber, C.; Magg, S.; and Wermter, S.
\newblock 2018.
\newblock A context-based approach for dialogue act recognition using simple
  recurrent neural networks.
\newblock {\em CoRR} abs/1805.06280.

\bibitem[\protect\citeauthoryear{Chen \bgroup et al\mbox.\egroup
  }{2018}]{LSTM_CRF}
Chen, Z.; Yang, R.; Zhao, Z.; Cai, D.; and He, X.
\newblock 2018.
\newblock Dialogue act recognition via crf-attentive structured network.
\newblock In {\em The 41st International ACM SIGIR Conference on Research \&
  Development in Information Retrieval},  225--234.
\newblock ACM.

\bibitem[\protect\citeauthoryear{Cho \bgroup et al\mbox.\egroup
  }{2014}]{grupaper}
Cho, K.; van Merrienboer, B.; G{\"{u}}l{\c{c}}ehre, {\c{C}}.; Bougares, F.;
  Schwenk, H.; and Bengio, Y.
\newblock 2014.
\newblock Learning phrase representations using {RNN} encoder-decoder for
  statistical machine translation.
\newblock {\em CoRR} abs/1406.1078.

\bibitem[\protect\citeauthoryear{Gimpel \bgroup et al\mbox.\egroup
  }{2013}]{diversity_bea_search}
Gimpel, K.; Batra, D.; Dyer, C.; and Shakhnarovich, G.
\newblock 2013.
\newblock A systematic exploration of diversity in machine translation.
\newblock In {\em Proceedings of EMNLP 2013},  1100--1111.

\bibitem[\protect\citeauthoryear{Janin \bgroup et al\mbox.\egroup
  }{2003}]{mrda}
Janin, A.; Baron, D.; Edwards, J.; Ellis, D.; Gelbart, D.; Morgan, N.; Peskin,
  B.; Pfau, T.; Shriberg, E.; Stolcke, A.; et~al.
\newblock 2003.
\newblock The icsi meeting corpus.
\newblock In {\em 2003 IEEE International Conference on Acoustics, Speech, and
  Signal Processing, 2003. Proceedings.(ICASSP'03).}, volume~1,  I--I.
\newblock IEEE.

\bibitem[\protect\citeauthoryear{Jozefowicz, Zaremba, and
  Sutskever}{2015}]{lstm_gru}
Jozefowicz, R.; Zaremba, W.; and Sutskever, I.
\newblock 2015.
\newblock An empirical exploration of recurrent network architectures.
\newblock In {\em International Conference on Machine Learning},  2342--2350.

\bibitem[\protect\citeauthoryear{Keizer, op~den Akker, and
  Nijholt}{2002}]{bayesian_dialog}
Keizer, S.; op~den Akker, R.; and Nijholt, A.
\newblock 2002.
\newblock Dialogue act recognition with bayesian networks for dutch dialogues.
\newblock In {\em Proceedings of the Third SIGdial Workshop on Discourse and
  Dialogue}.

\bibitem[\protect\citeauthoryear{Khanpour, Guntakandla, and
  Nielsen}{2016}]{LSTM_softmax}
Khanpour, H.; Guntakandla, N.; and Nielsen, R.
\newblock 2016.
\newblock Dialogue act classification in domain-independent conversations using
  a deep recurrent neural network.
\newblock In {\em COLING}.

\bibitem[\protect\citeauthoryear{Kingma and Ba}{2014}]{adam}
Kingma, D.~P., and Ba, J.
\newblock 2014.
\newblock Adam: A method for stochastic optimization.
\newblock {\em arXiv preprint arXiv:1412.6980}.

\bibitem[\protect\citeauthoryear{Kumar \bgroup et al\mbox.\egroup
  }{2018}]{hierarchical_LSTM_CRF}
Kumar, H.; Agarwal, A.; Dasgupta, R.; and Joshi, S.
\newblock 2018.
\newblock Dialogue act sequence labeling using hierarchical encoder with crf.
\newblock In {\em Thirty-Second AAAI Conference on Artificial Intelligence}.

\bibitem[\protect\citeauthoryear{LeCun, Bengio, and Hinton}{2015}]{dropout}
LeCun, Y.; Bengio, Y.; and Hinton, G.
\newblock 2015.
\newblock Deep learning.
\newblock {\em nature} 521(7553):436.

\bibitem[\protect\citeauthoryear{Li \bgroup et al\mbox.\egroup }{2016a}]{mmi}
Li, J.; Galley, M.; Brockett, C.; Gao, J.; and Dolan, W.~B.
\newblock 2016a.
\newblock A diversity-promoting objective function for neural conversation
  models.
\newblock In {\em HLT-NAACL}.

\bibitem[\protect\citeauthoryear{Li \bgroup et al\mbox.\egroup
  }{2016b}]{persona_based}
Li, J.; Galley, M.; Brockett, C.; Spithourakis, G.; Gao, J.; and Dolan, B.
\newblock 2016b.
\newblock A persona-based neural conversation model.
\newblock In {\em Proceedings of the 54th Annual Meeting of ACL (Volume 1: Long
  Papers)},  994--1003.
\newblock Association for Computational Linguistics.

\bibitem[\protect\citeauthoryear{Li \bgroup et al\mbox.\egroup
  }{2018a}]{crf_multi_task}
Li, R.; Lin, C.; Collinson, M.; Li, X.; and Chen, G.
\newblock 2018a.
\newblock A dual-attention hierarchical recurrent neural network for dialogue
  act classification.
\newblock {\em CoRR}.

\bibitem[\protect\citeauthoryear{Li \bgroup et al\mbox.\egroup
  }{2018b}]{dependancy}
Li, Z.; Cai, J.; He, S.; and Zhao, H.
\newblock 2018b.
\newblock Seq2seq dependency parsing.
\newblock In {\em Proceedings of the 27th International Conference on
  Computational Linguistics},  3203--3214.

\bibitem[\protect\citeauthoryear{Loshchilov and Hutter}{2017}]{adamw}
Loshchilov, I., and Hutter, F.
\newblock 2017.
\newblock Fixing weight decay regularization in adam.
\newblock {\em arXiv preprint arXiv:1711.05101}.

\bibitem[\protect\citeauthoryear{Luong, Pham, and Manning}{2015}]{attention}
Luong, T.; Pham, H.; and Manning, C.~D.
\newblock 2015.
\newblock Effective approaches to attention-based neural machine translation.
\newblock In {\em Proceedings of EMNLP 2015},  1412--1421.
\newblock Lisbon, Portugal: Association for Computational Linguistics.

\bibitem[\protect\citeauthoryear{Mikolov \bgroup et al\mbox.\egroup
  }{2013}]{word2vec}
Mikolov, T.; Sutskever, I.; Chen, K.; Corrado, G.~S.; and Dean, J.
\newblock 2013.
\newblock Distributed representations of words and phrases and their
  compositionality.
\newblock In {\em NeurIPS},  3111--3119.

\bibitem[\protect\citeauthoryear{Pennington, Socher, and Manning}{2014}]{glove}
Pennington, J.; Socher, R.; and Manning, C.
\newblock 2014.
\newblock Glove: Global vectors for word representation.
\newblock In {\em Proceedings of EMNLP 2014},  1532--1543.

\bibitem[\protect\citeauthoryear{Raheja and Tetreault}{2019}]{DAClassifCxt}
Raheja, and Tetreault.
\newblock 2019.
\newblock Dialogue act classification with context-aware self-attention.
\newblock {\em CoRR} abs/1904.02594.

\bibitem[\protect\citeauthoryear{Shriberg \bgroup et al\mbox.\egroup
  }{2004}]{mrda_2}
Shriberg, E.; Dhillon, R.; Bhagat, S.; Ang, J.; and Carvey, H.
\newblock 2004.
\newblock The icsi meeting recorder dialog act (mrda) corpus.
\newblock In {\em Proceedings of the 5th SIGdial Workshop on Discourse and
  Dialogue at HLT-NAACL 2004}.

\bibitem[\protect\citeauthoryear{Sordoni \bgroup et al\mbox.\egroup
  }{2015}]{hierarchical_encoder}
Sordoni, A.; Bengio, Y.; Vahabi, H.; Lioma, C.; Grue~Simonsen, J.; and Nie,
  J.-Y.
\newblock 2015.
\newblock A hierarchical recurrent encoder-decoder for generative context-aware
  query suggestion.
\newblock In {\em Proceedings of the 24th ACM International on Conference on
  Information and Knowledge Management},  553--562.
\newblock ACM.

\bibitem[\protect\citeauthoryear{Stolcke \bgroup et al\mbox.\egroup
  }{1998}]{switchboard_da}
Stolcke, A.; Shriberg, E.; Bates, R.; Coccaro, N.; Jurafsky, D.; Martin, R.;
  Meteer, M.; Ries, K.; Taylor, P.; Van Ess-Dykema, C.; et~al.
\newblock 1998.
\newblock Dialog act modeling for conversational speech.
\newblock In {\em AAAI Spring Symposium on Applying Machine Learning to
  Discourse Processing},  98--105.

\bibitem[\protect\citeauthoryear{Stolcke \bgroup et al\mbox.\egroup
  }{2000}]{handcrafted_HMM}
Stolcke, A.; Ries, K.; Coccaro, N.; Shriberg, E.; Bates, R.; Jurafsky, D.;
  Taylor, P.; Martin, R.; Ess-Dykema, C.~V.; and Meteer, M.
\newblock 2000.
\newblock Dialogue act modeling for automatic tagging and recognition of
  conversational speech.
\newblock {\em Computational linguistics} 26(3):339--373.

\bibitem[\protect\citeauthoryear{Surendran and Levow}{2006}]{svm_dialog}
Surendran, D., and Levow, G.-A.
\newblock 2006.
\newblock Dialog act tagging with support vector machines and hidden markov
  models.
\newblock In {\em Ninth International Conference on Spoken Language
  Processing}.

\bibitem[\protect\citeauthoryear{Sutskever, Vinyals, and Le}{2014}]{seq2seq}
Sutskever, I.; Vinyals, O.; and Le, Q.~V.
\newblock 2014.
\newblock Sequence to sequence learning with neural networks.
\newblock In {\em NeurIPS},  3104--3112.

\bibitem[\protect\citeauthoryear{Wiseman and
  Rush}{2016}]{beam_seach_optimization}
Wiseman, S., and Rush, A.~M.
\newblock 2016.
\newblock Sequence-to-sequence learning as beam-search optimization.
\newblock In {\em EMNLP}.

\bibitem[\protect\citeauthoryear{Wu \bgroup et al\mbox.\egroup
  }{2016}]{beam_search_alpha}
Wu, Y.; Schuster, M.; Chen, Z.; Le, Q.~V.; Norouzi, M.; Macherey, W.; Krikun,
  M.; Cao, Y.; Gao, Q.; Macherey, K.; et~al.
\newblock 2016.
\newblock Google's neural machine translation system: Bridging the gap between
  human and machine translation.
\newblock {\em arXiv preprint arXiv:1609.08144}.

\end{thebibliography}
\bibliographystyle{aaai}

\appendix
\clearpage
\appendix\section{Appendix}
\label{sec:appendix}

\subsection{Additional details on the datasets}
\begin{table}[ht!]
\begin{center}
\resizebox{\columnwidth}{!}{%
    \begin{tabular}{c|c|c|c|c|c}
        \hline
        Dataset & $|C|$ & $|V|$ & Train & Val & Test \\
        \hline
        MRDA & 5 & 10K & 51(76K) & 11(15K) & 11(15K) \\
        \hline
        SwDA & 42 & 19K & 1003(173K) & 112(22K)& 19(9K)  \\
        \hline
    \end{tabular}
    }
    \caption{Statistics for MRDA and SwDA. $|C|$ is the number of Dialogue Act classes, $|V|$ is the vocabulary size. Training, Validation and Testing indicate the number of conversations (number of utterances) in the respective splits.}
\end{center}
\end{table}
\textbf{Tags in SwDA}:
SwDA extends the Switchboard-1 corpus with tags from the SWBD-DAMSL tagset. The 220 tags were reduced to 42 tags. The resulting tags include dialogue acts like statement-non-opinion, acknowledge, statement-opinion, agree/accept, etc. The average speaker turns per conversation, tokens per conversation, and tokens per utterance are 195.2, 1,237.8, and 7.0, respectively.

\subsection{Full results for Experiment 4: How to leverage beam search to improve the performance?}
Tab \ref{tab:f1_all_all} shows the influence of the varying number of beam size during inference.
\begin{table*}[!htb]

\centering
\begin{tabular}{c|ccc|ccc|ccc|ccc}
\hline 
&\multicolumn{12}{c}{SwDA} \\ \hline
     \diagbox[height=20pt]{\footnotesize{Encoder}}{\footnotesize{Decoder}}  & \multicolumn{3}{c}{$\text{VGRU}_\text{D}$} & \multicolumn{3}{c}{GRU att.} & \multicolumn{3}{c}{GRU soft guid. att.} & \multicolumn{3}{c}{GRU hard guid. att.}\\ \hline
     Beam Size &  1 & 2 & 5   &1 & 2 & 5 & 1 & 2 & 5 &1 & 2 & 5\\ \hline

  {$\text{VGRU}_\text{E}$}  & 81.6&82.0&81.9     &82.0&82.0&82.1  &   82.8&83.0&82.9   &82.9&83.0&83.0 \\ 
  {HGRU} & 82.4&82.3&82.2     &82.3&82.2&82.2    &   83.1&83.8&83.8 &84.0&84.4&84.4 \\ 
  {PersoHGRU}  & 49.8&49.8&50.0     &79.4&79.1&78.9    &   84.0&84.0&84.0 &83.5&83.6&83.5 \\ \hline
    
&\multicolumn{12}{c}{MRDA} \\\hline
      \diagbox[height=20pt]{\footnotesize{Encoder}}{\footnotesize{Decoder}} & \multicolumn{3}{c}{$\text{VGRU}_\text{D}$} & \multicolumn{3}{c}{GRU att.} & \multicolumn{3}{c}{GRU soft guid. att.} & \multicolumn{3}{c}{GRU hard guid. att.} \\ \hline
     Beam Size &  1 & 2 & 5   &1 & 2 & 5 & 1 & 2 & 5 &1 & 2 & 5\\ \hline

  {$\text{VGRU}_\text{E}$}  & 88.5&88.5&88.5     & 88.5&88.5&88.5 &88.2&88.3&88.3 &88 .7&88.8&88.8   \\ 
  {HGRU} & 90.0&90.0&90.0     &89.9&90.0&90.3     &   90.0&90.2&90.2 &90.4&90.4&90.4 \\ 
  {PersoHGRU} & 66.2&66.2&66.5     &87.7&87.7&87.8     &   88.2&88.7&88.9  &86.9&86.9&86.9     \\ 
   
     \hline
\end{tabular}
\caption{Accuracy on the dev set of the different encoder/decoder combination  MRDA and SwDA.}
\label{tab:f1_all_all}
\end{table*}

\subsection{Error analysis}
\begin{figure}[htb!]
	   \center{\includegraphics[width=0.4\textwidth]
	       {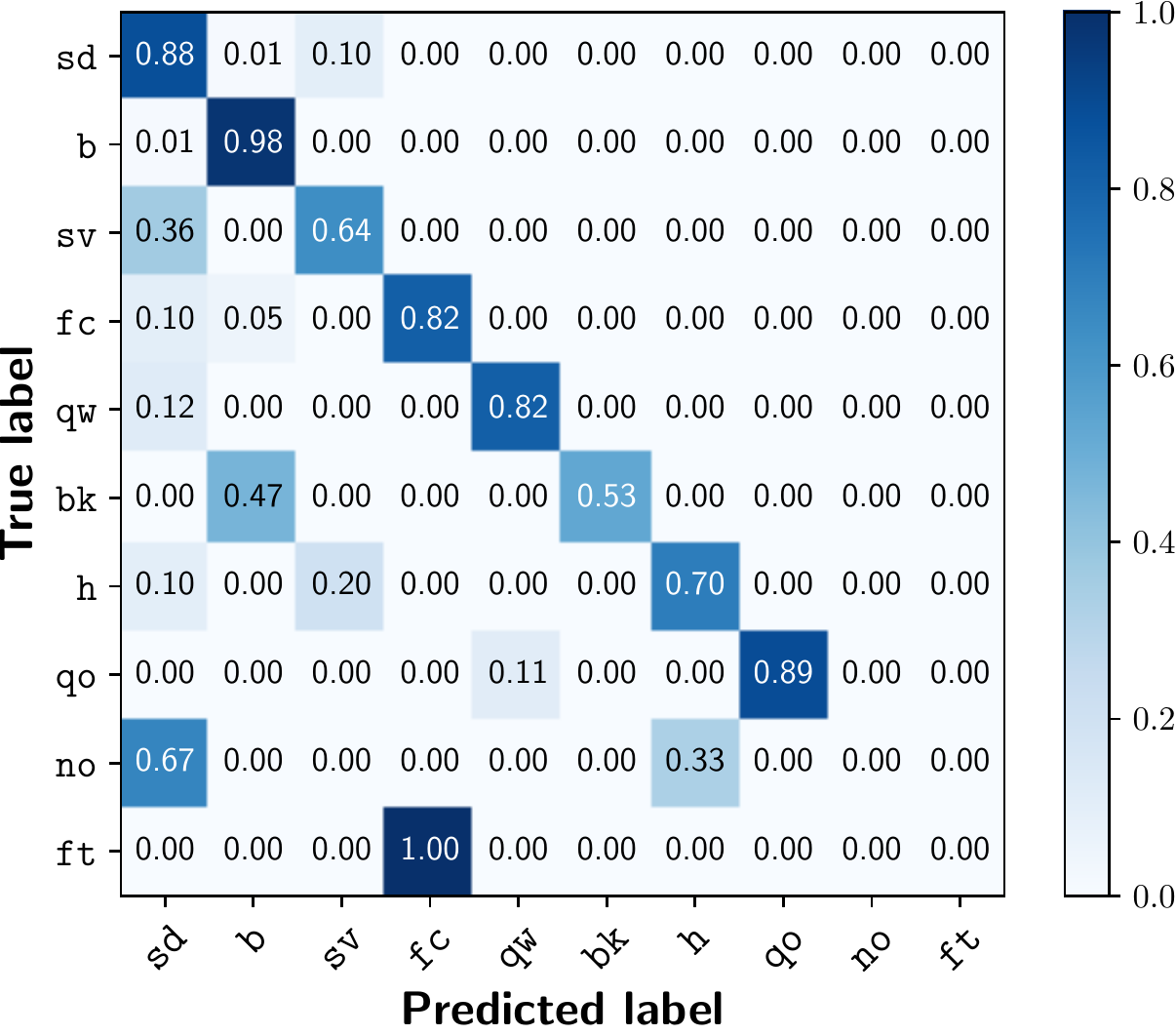}}
	  \caption{\label{fig:confusion-matrix} Confusion Matrix for our best performing seq2seq model on SwDA for 10 out of 42 tags. For label designation see Section~\ref{ssec:datasets}.}
\end{figure}

The confusion matrix on SwDA (see Figure~\ref{fig:confusion-matrix}) illustrates that our model faces same difficulties as human annotator: \texttt{sd} is often confused with \texttt{sv}, \texttt{bk} with \texttt{b}, \texttt{qo} with \texttt{qw}. Due to high imbalance of SwDA, our system fails to recognise underrepresented labels (e.g. \texttt{no} and \texttt{ft}).

The confusion Matrix on MRDA shows that, here, the DA classification is easier compared to SwDA with fewer tags and classes that are more easily distinguished. $seq2seq_{BEST}$ reaches a perfect score at recognising questions. One of the reasons for the mislabelling between backchannel (\texttt{b}) and statement (\texttt{s}) is that the MRDA dataset is highly imbalanced, with more than 50\% of the utterances labelled as class \texttt{s}.

\begin{figure}[htb!]
  \center{\includegraphics[width=0.50\textwidth]
      {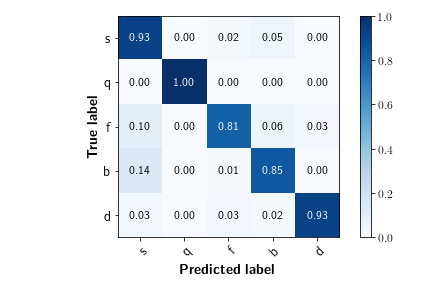}}
  \caption{\label{fig:confusion-matrix_2} Confusion Matrix for our best performing seq2seq model on MRDA. For label designation see Section~\ref{ssec:datasets}.}
\end{figure}

%\begin{figure}[!h]

% OPTIMIZER ETC

% \begin{figure}
% 	   \center{\includegraphics[width=0.45\textwidth]
% 	       {images/decoder_hidden_2.png}}
% 	       \label{fig:hidden}

% 	\end{figure}

% \subsection{Decoder Size}
% In our work we have presented a seq2seq architecture that achieves the highest accuracy for DA classification on SwDA and MRDA. The best performing model has been deeply inspired by NMT approaches.In NMT, a complex decoder is required to produce a correct sentence. In DA classification, given the limited size of the output space, the decoder can be much simpler. We propose to reduce the impact of the decoder hidden size on the performances.

% Figure~\ref{fig:hidden} show different accuracy score for the hierarchical encoder (fixed hidden size of 128) with \textit{hard guided attention} for different decoder hidden size. We observe that the decoder with restraint size exhibits high accuracy score with fewer parameters (e.g, a decoder with hidden size 8 leads to an accuracy of 80\%). We note that increasing the decoder hidden size to more than 48 does not really improve the results.

% \appendix

% \section{Appendices}
% \label{sec:appendix}
% Appendices are material that can be read, and include lemmas, formulas, proofs, and tables that are not critical to the reading and understanding of the paper. 
% Appendices should be \textbf{uploaded as supplementary material} when submitting the paper for review. Upon acceptance, the appendices come after the references, as shown here. Use
% \verb|\appendix| before any appendix section to switch the section
% numbering over to letters.

\end{document}